\documentclass[conference]{IEEEtran}
\IEEEoverridecommandlockouts
\usepackage{cite}
\usepackage{amsmath,amssymb,amsfonts}
\usepackage{algorithmic}
\usepackage{graphicx}
\usepackage{textcomp}
\usepackage{xcolor}
\usepackage{booktabs}
\usepackage{multirow}
\usepackage{url}

\def\BibTeX{{\rm B\kern-.05em{\sc i\kern-.025em b}\kern-.08em
    T\kern-.1667em\lower.7ex\hbox{E}\kern-.125emX}}
\begin{document}

\title{DriftGuard: Safety-Aware Multi-Monitor Detection and Selective Adaptation for Evolving Toxicity Moderation}

\author{

\IEEEauthorblockN{
\begin{tabular}{c c}
\begin{tabular}{c}
Yuting Xin\textsuperscript{$\dagger$*}\\
\textit{Department of Information}\\
\textit{and Decision Sciences}\\
\textit{University of Minnesota}\\
Minneapolis, USA\\
yuting.xin@outlook.com
\end{tabular}
&
\begin{tabular}{c}
Hanyu Cai{$\dagger$}\\
\textit{Department of Industrial Engineering}\\
\textit{and Management Sciences}\\
\textit{Northwestern University}\\
Evanston, USA\\
hanyucai2022@u.northwestern.edu
\end{tabular}
\end{tabular}
}

\vspace{1.5em}

\IEEEauthorblockN{
\begin{tabular}{c c c}
\begin{tabular}{c}
Binqi Shen\\
\textit{Department of Industrial Engineering}\\
\textit{and Management Sciences}\\
\textit{Northwestern University}\\
Evanston, USA\\
binqishen2021@u.northwestern.edu
\end{tabular}
&
\begin{tabular}{c}
Lier Jin\\
\textit{Fuqua School of Business}\\ 
\textit{Duke University}\\
Durham, USA\\
lierjin@alumni.duke.edu
\end{tabular}
&
\begin{tabular}{c}
Lan Hu\\
\textit{Department of Engineering}\\
\textit{Carnegie Mellon University}\\
Pittsburgh, USA\\
lanh@alumni.cmu.edu
\end{tabular}
\end{tabular}
}

\vspace{1.5em}

\IEEEauthorblockA{
\vspace{0.8em}
\textsuperscript{$\dagger$}Equal contribution \quad
\textsuperscript{*}Corresponding author
}
}

\maketitle

\begin{abstract}
Automated toxicity moderation systems operate in dynamic online environments where harmful behavior evolves through coded language, shifting targets, and strategic adaptation to enforcement. Existing drift detection methods often focus on global distributional change, but such signals may miss safety-relevant shifts that emerge in localized harm subspaces or high-risk model-error regions. This paper introduces \textit{DriftGuard}, a safety-aware adaptive moderation framework that combines multi-monitor drift detection with selective model updating. The framework tracks global text drift, identity-harm drift, model uncertainty, toxic-risk drift, and false-negative-risk drift. When safety-relevant change is detected, the model is updated using a hard-mix adaptation set that prioritizes likely false negatives, identity-related high-risk examples, false-positive-risk examples, and uncertain boundary cases. Experiments on Civil Comments temporal shift and Jigsaw-to-DynaHate cross-dataset shift show that safety-aware monitors detect risks missed by global drift alone. Hard-mix adaptation improves toxic recall and accuracy over no-update and random-balanced baselines. It raises toxic recall from 0.8453 to 0.8777 on Civil Comments and from 0.7107 to 0.8523 on DynaHate. Bootstrap analysis further shows stable DynaHate safety gains, with toxic recall increasing by 0.1418 and false-negative prevalence decreasing by 0.0781. Overall, DriftGuard links safety-aware drift detection with targeted, lightweight model updating for more robust adaptive toxicity moderation.
\end{abstract}

\begin{IEEEkeywords}
Toxicity Moderation, Hate Speech Detection, Distribution Shift, Concept Drift, Drift Detection, Safety-Aware Monitoring, Adaptive Moderation, Selective Adaptation, Hard Example Mining, Parameter-Efficient Fine-Tuning
\end{IEEEkeywords}

\section{Introduction}

Automated toxicity moderation systems are increasingly deployed to identify harmful user-generated content at large scale. Although modern classifiers can perform well on their training distributions, moderation environments change over time. Users adopt coded language, shift targets, and strategically evade enforcement, including through adversarial prompts and multi-agent mediation patterns \cite{gama2014survey,yao2026rankingabusestrategicpairwise,lin2026reflect,li2026dangerousobjectiveoppositeadvice}. As a result, a model that initially detects toxic or hateful content reliably may become less effective as harmful behavior evolves. This creates a practical need for moderation systems that can detect safety-relevant change and adapt efficiently without relying on frequent full retraining\cite{rabanser2019failing}.

A central challenge is that distribution shift in moderation is not always visible at the aggregate level\cite{borkan2019nuanced,sap2019risk}. Recent work on LLM character understanding shows that apparent benchmark success can reflect memorization rather than genuine reasoning. This motivates monitoring safety-relevant behavior rather than relying only on aggregate performance signals \cite{jiang2026beyond}. Many drift detection methods track broad input or prediction changes. However, harmful behavior can emerge in localized safety-critical regions, so moderation systems need safety-relevant monitoring beyond aggregate metrics \cite{wang2026safeskillscollidemeasuring}. In these cases, global text drift may remain modest even while the model becomes more likely to miss harmful content\cite{polo2023unified}. Therefore, adaptive moderation requires monitoring signals that are aligned not only with distributional change, but also with moderation risk\cite{rottger2021hatecheck,wang2026safeskillscollidemeasuring,qian2026relevantwarrantedevidenceforcecalibration}.

This paper proposes a safety-aware adaptive moderation framework that links multi-monitor drift detection with selective model updating. The framework monitors incoming data using global distribution drift, harm-subspace drift, model uncertainty, toxic-risk drift, and false-negative-risk drift. When one or more monitors indicate safety-relevant change, the model is updated using a hard-mix adaptation set composed of high-risk and informative examples, rather than a random sample of recent data. This design allows adaptation to focus on the types of examples most likely to affect moderation safety while keeping the update targeted and lightweight\cite{settles2009active}.

The paper makes three contributions. First, it formulates moderation drift as a safety-aware monitoring problem in which global distribution shift is supplemented by harm-subspace and model-risk signals. Second, it introduces a multi-monitor trigger mechanism paired with hard-mix selective adaptation, connecting drift detection directly to model updating. Third, it evaluates the framework under temporal and cross-domain moderation shift, showing that safety-aware monitors can detect targeted harm shifts missed by global drift alone and that hard-mix adaptation improves toxic recall and overall robustness compared with no-update and random-balanced updating baselines.

\section{Related Work}

Prior work has studied drift detection, toxicity moderation, and efficient adaptation largely as separate problems. However, adaptive moderation requires these components to be connected. The system must detect safety-relevant drift, determine whether the drift affects harmful or high-risk subspaces, and update the model using informative examples rather than random recent data. Our work addresses this gap through a multi-monitor harm-aware trigger and a hard-mix adaptation strategy.

\subsection{Distribution Shift and Drift Monitoring}

Distribution shift is a central challenge for deployed machine learning systems. A model is typically trained under one data-generating distribution but deployed under conditions that may change over time, producing covariate shift, label shift, or concept drift. Prior work has formalized these shift types and developed methods for detecting and adapting to non-stationary data streams. Surveys on concept drift emphasize that deployed models require not only drift detection, but also drift understanding and adaptation, because unaddressed drift can lead to model degradation over time \cite{bayram2022concept}. More recent deployment-focused work similarly argues that model monitoring should track features, predictions, uncertainty, performance-related signals, and explanation stability rather than assuming that a static validation set remains representative after deployment \cite{polo2023unified,lin2025shap,fu2026sphunchypersphericaluncertaintydecomposition}.

A large body of work detects drift by comparing source and target distributions using statistical tests, divergence measures, or learned representations. Rabanser et al. show that dataset shift can often be detected through two-sample testing after dimensionality reduction, and that domain-discriminating models can help characterize whether observed shift is harmful \cite{rabanser2019failing}. Other work focuses on specific forms of shift, such as label shift, where the class prior changes while the class-conditional distribution remains stable. For example, Black Box Shift Estimation uses a trained classifier's predictions to estimate and correct label shift without requiring target labels \cite{lipton2018detecting}. Benchmarks such as WILDS further show that real-world distribution shifts can substantially reduce out-of-distribution performance even when models perform well in-distribution \cite{koh2021wilds}.

However, existing drift monitoring methods often treat drift as a global property of the input or prediction distribution. This is limiting for safety-critical moderation. First, statistically detectable shift does not always imply safety-relevant degradation. Feature or prediction drift can occur without performance loss, so monitoring should check model behavior rather than raw outputs alone \cite{eck2022monitoring,chen2026learning}. Recent RAG reliability studies provide important evidence that surface-level relevance is not sufficient for reliable model behavior: retrieved context can shape outputs under knowledge conflict, topically relevant citations may still fail to warrant the generated claim, and routing choices can affect robustness--precision tradeoffs \cite{chen2026rag,qian2026relevantwarrantedevidenceforcecalibration,cheng2026sustainablehybriddocumentroutedretrieval}. Second, global drift metrics may obscure localized changes in subpopulations or high-risk regions of the data. In moderation, a small increase in identity-targeted abuse or false-negative-risk examples may be operationally important even if the aggregate input distribution changes only modestly. Related work on discrepancy-aware fusion and structured semantic signals also highlights the value of integrating complementary signals under noisy or domain-specific conditions \cite{dang2025discrepancy,zhang2026finsentllm}.
Our work builds on drift monitoring research but shifts the focus from global distribution change alone to safety-aware monitoring, combining global drift with harm-subspace and model-risk signals.

\subsection{Toxicity Moderation and Harm Subspaces}

Automated toxicity and hate speech detection is challenging because harmful content varies by target group, linguistic form, and social context. Prior work shows that aggregate classification metrics can hide important subgroup failures. Systematic reviews of hate speech detection identify persistent issues including ambiguous task definitions, class imbalance, contextual dependence, and limited cross-domain generalization \cite{jahan2023systematic, pavlopoulos2020toxicity}.

A key concern is unintended bias in moderation models. Borkan et al. introduce Civil Comments identity annotations and metrics for measuring subgroup bias in toxicity classifiers \cite{borkan2019nuanced}. Abusive-language classifiers can over-predict toxicity for dialectal or identity-associated language, causing disproportionate errors for marginalized groups. Causal debiasing work further motivates robustness to spurious shortcuts in pretrained language models \cite{sap2019risk,davidson2019racial,zhou2023causal}. These findings motivate evaluation and monitoring at the harm-subspace level rather than only at the aggregate dataset level. Related work on fairness in synthetic medical data similarly shows that subgroup representation imbalances can persist in generated datasets, reinforcing the need to monitor protected or high-risk subspaces rather than relying only on aggregate data quality \cite{salarian2025medequalizer}.

Recent benchmarks further emphasize targeted diagnostic evaluation. HateXplain provides target-community and rationale annotations \cite{mathew2021hatexplain}, HateCheck introduces functional tests for hate speech models \cite{rottger2021hatecheck}, and DynaHate uses human-and-model-in-the-loop data generation to expose challenging hate speech examples \cite{vidgen2021learning}. These studies show that moderation failures often appear in specific behavioral or identity-related regions of the data.

Our work builds on this literature by moving harm-subspace signals from post-hoc diagnostic evaluation to drift-triggered adaptation. Instead of monitoring only global text drift, our framework tracks safety-relevant subspaces, such as identity-harm and false-negative-risk regions, and uses these signals to decide when model updating is needed.

\subsection{Selective Adaptation and Efficient Model Updating}

A separate line of work studies how models can be updated efficiently when new data becomes available. Active learning selects informative examples for annotation or training, often using uncertainty, diversity, or expected model improvement as selection criteria \cite{zhang2022survey, li2024survey}. Related hard-example mining, cost-sensitive learning, and imbalanced-classification methods prioritize difficult, high-risk, or underrepresented examples rather than treating all training samples equally \cite{wu2026classweightingversusconditioning,han2026validationstagecombinatorialfusionanalysis}. Online hard example mining and focal loss both show that emphasizing hard or high-loss examples can improve learning efficiency under class imbalance or many easy examples \cite{shrivastava2016training, lin2017focal}. These ideas motivate selective updating strategies that focus adaptation on examples most likely to affect model behavior\cite{xu2026paceddistillationonpolicyselfdistillation}. Recent work on on-policy distillation introduces token-importance selection, showing that high-entropy and high-divergence tokens can provide especially useful learning signals and can be prioritized to reduce training cost while preserving performance \cite{xu2026tiptokenimportanceonpolicy}.

For large neural models, full fine-tuning can be computationally expensive and impractical for frequent updates. Parameter-efficient fine-tuning methods address this by updating only a small subset of parameters or adding lightweight trainable modules. LoRA freezes the base model and learns low-rank update matrices, substantially reducing trainable parameters and memory cost while preserving downstream performance \cite{hu2022lora}. Recent PEFT surveys further show that parameter-efficient adaptation has become a practical approach for customizing large models under limited compute and deployment constraints \cite{han2024parameter, wang2024parameter,a8,zhou2025gsq}. Related work on structured medical report normalization likewise highlights the value of transforming noisy textual inputs into more consistent supervision for robust model training \cite{chu2026medtri}. This emphasis on deployment efficiency is consistent with recent work on large reasoning models, where pruning and distillation have been used to reduce reasoning cost while preserving task performance \cite{jiang2026drpdistilledreasoningpruning}. Related work on adaptive distributed learning and sustainable on-device intelligence similarly highlights the value of monitoring deployment conditions and balancing efficiency with model performance \cite{10.1007/978-3-031-99872-0_20,cheng2026toward}. Related work on cross-attention token pruning also shows that identifying and removing less informative tokens can improve inference efficiency while preserving accuracy \cite{11050741}.

Our work connects selective sample choice with parameter-efficient updating for adaptive moderation. Rather than updating on random recent data, the proposed hard-mix strategy selects high-risk and high-uncertainty examples associated with moderation failures, including likely false negatives, identity-related high-risk examples, false-positive-risk examples, and uncertain boundary cases. Related work on prompt compression and cross-attention token pruning also shows that identifying and removing less informative tokens can improve inference efficiency while preserving accuracy \cite{11460600,11050741}. Related work on prompt indicators further suggests that lightweight input-design choices can affect LLM behavior, reinforcing the need for careful evaluation when deploying adaptive language systems \cite{a6}. This differs from general active learning or PEFT work by using safety-aware drift monitors to decide both when adaptation should be triggered and which examples should drive the update.

\section{Methodology}

\subsection{Framework Overview}

This paper studies adaptive moderation under safety-relevant distribution shift. The proposed framework makes two linked decisions. First, it monitors incoming data using multiple drift and model-risk signals. Second, when one or more monitors indicate safety-relevant shift, it updates the model using a selectively constructed hard-mix adaptation set rather than a random sample of recent data or a full retraining corpus.

The framework is motivated by the fact that harmful behavior can change in localized subspaces. Identity-targeted abuse, coded harmful language, uncertain boundary cases, or likely false negatives may increase even when the aggregate text distribution changes only modestly. Therefore, the framework combines broad distribution monitoring with harm-subspace and model-risk monitoring, then uses the detected failure modes to guide model updating.

\subsection{Task Definition}

We formulate moderation as binary toxic-content classification. Given a data sample \(x\), the model predicts

\[
f(x) \rightarrow \{\text{non-toxic}, \text{toxic}\}.
\]

The classifier also outputs a toxic-class probability \(p_{\mathrm{toxic}}(x)\). This probability is used for classification, threshold calibration, uncertainty monitoring, toxic-risk monitoring, and adaptation-sample ranking. The primary safety objective is to preserve recall on toxic content under shift, because false negatives correspond to harmful comments that remain undetected.

\subsection{Multi-Monitor Drift Detection}

At each monitoring interval, the framework compares a target batch \(B_t\) with a reference batch \(B_{\mathrm{ref}}\), where \(B_{\mathrm{ref}}\) represents the model's baseline operating distribution. The trigger uses five monitor families: global distribution drift, identity-harm drift, model uncertainty drift, toxic-risk drift, and false-negative-risk drift.

\subsubsection{Global Distribution Drift}

Global drift measures broad text-distribution shift. Comments are represented with hashed character \(n\)-gram features, and Jensen-Shannon distance is computed between the reference and target feature distributions:

\[
D_{\mathrm{global}} =
JS\!\left(P_{\mathrm{ngram}}(B_{\mathrm{ref}}),
P_{\mathrm{ngram}}(B_t)\right).
\]

This monitor captures major changes in language distribution, but it does not determine whether the shift is safety-relevant.

\subsubsection{Identity-Harm Drift}

Identity-harm drift measures changes in identity-targeted harmful content when identity annotations are available. For example, this monitor uses the \texttt{identity\_attack} annotation in the Civil Comments dataset. We compute both the change in identity-attack rate and the change in mean identity-attack score:

\[
D_{\mathrm{identity\_rate}} =
\mathrm{rate}_{\mathrm{identity}}(B_t)
-
\mathrm{rate}_{\mathrm{identity}}(B_{\mathrm{ref}}),
\]

\[
D_{\mathrm{identity\_score}} =
\mathrm{mean}_{\mathrm{identity}}(B_t)
-
\mathrm{mean}_{\mathrm{identity}}(B_{\mathrm{ref}}).
\]

This monitor captures localized harm shifts that may be diluted in aggregate text statistics.

\subsubsection{Model Uncertainty Drift}

Uncertainty drift measures whether the model becomes less confident on the target batch. For each data sample, predictive entropy is computed from the toxic probability:

\[
H(x) =
-p_{\mathrm{toxic}}(x)\log_2 p_{\mathrm{toxic}}(x)
-
\left(1-p_{\mathrm{toxic}}(x)\right)
\log_2\left(1-p_{\mathrm{toxic}}(x)\right).
\]

The uncertainty monitor is the change in mean entropy:

\[
D_{\mathrm{entropy}} =
\frac{1}{|B_t|}\sum_{x \in B_t}H(x)
-
\frac{1}{|B_{\mathrm{ref}}|}\sum_{x \in B_{\mathrm{ref}}}H(x).
\]

An increase indicates that the target batch contains more examples which are close to the model's decision boundary.

\subsubsection{Toxic-Risk Drift}

Toxic-risk drift measures whether the model assigns greater toxic risk to the target batch. We use two quantities:

\[
D_{\mathrm{toxic\_prob}} =
\frac{1}{|B_t|}\sum_{x \in B_t}p_{\mathrm{toxic}}(x)
-
\frac{1}{|B_{\mathrm{ref}}|}\sum_{x \in B_{\mathrm{ref}}}p_{\mathrm{toxic}}(x),
\]

\[
D_{\mathrm{toxic\_prev}} =
\mathrm{PredToxicRate}(B_t)
-
\mathrm{PredToxicRate}(B_{\mathrm{ref}}).
\]

The first monitor captures changes in mean toxic probability, while the second captures changes in predicted toxic prevalence after thresholding.

\subsubsection{False-Negative-Risk Drift}

False-negative-risk drift measures whether the model becomes more likely to miss toxic content. In the offline experiments, ground-truth toxicity labels are available, allowing us to directly identify toxic examples that the model predicts as non-toxic. We compute the aggregate false-negative prevalence of a batch as

\[
\mathrm{FNPrev}(B) =
\frac{\sum_{x \in B}\mathbb{1}[y(x)=\text{toxic} \land \hat{y}(x)=\text{non-toxic}]}
{|B|}.
\]

The false-negative-risk drift monitor is then

\[
D_{\mathrm{fn\_risk}} =
\mathrm{FNPrev}(B_t)
-
\mathrm{FNPrev}(B_{\mathrm{ref}}).
\]

This quantity differs from the label-conditional false-negative rate, which is \(1-\) toxic recall. We report toxic recall separately, and use aggregate false-negative prevalence to measure the share of all evaluated comments that are toxic but missed. In live deployment, immediate labels may be unavailable. This monitor can instead use delayed reviews, weak safety labels, model disagreement, high-risk lexical indicators, or policy-specific risk signals.

\subsection{Trigger Rule}

The framework triggers adaptation when any monitor exceeds its operating threshold:

\[
\begin{aligned}
\mathrm{trigger}(B_t) =
&[D_{\mathrm{global}} > \theta_{\mathrm{global}}] \\
&\lor [D_{\mathrm{identity\_rate}} > \theta_{\mathrm{identity\_rate}}] \\
&\lor [D_{\mathrm{identity\_score}} > \theta_{\mathrm{identity\_score}}] \\
&\lor [D_{\mathrm{entropy}} > \theta_{\mathrm{entropy}}] \\
&\lor [D_{\mathrm{toxic\_prob}} > \theta_{\mathrm{toxic\_prob}}] \\
&\lor [D_{\mathrm{toxic\_prev}} > \theta_{\mathrm{toxic\_prev}}] \\
&\lor [D_{\mathrm{fn\_risk}} > \theta_{\mathrm{fn\_risk}}].
\end{aligned}
\]

The rule is intentionally disjunctive because the monitors represent different failure modes. The framework therefore reports individual monitor values and trigger decisions rather than collapsing them into a single aggregate score.

The thresholds are monitor-specific operating points. The raw monitor values are not directly comparable because Jensen-Shannon distance measures feature-distribution divergence, while the other monitors measure changes in harm annotations or model behavior. The thresholds should therefore be interpreted as risk-tolerance settings for each monitor rather than universal drift magnitudes.

\subsection{Hard-Mix Selective Adaptation}

When the trigger detects safety-relevant shift, the model is updated using hard-mix selective adaptation. Given an adaptation budget \(K\), the method ranks recent examples into four priority groups. Table I summarizes the target allocation and ranking criterion for each component of the hard-mix adaptation set.

\begin{table}[t]
\centering
\caption{Hard-mix adaptation components.}
\label{tab:hard-mix-components}
\scriptsize
\setlength{\tabcolsep}{3pt}
\renewcommand{\arraystretch}{1.08}
\begin{tabular}{p{0.36\columnwidth} p{0.15\columnwidth} p{0.39\columnwidth}}
\toprule
Component & Share & Ranking criterion \\
\midrule
Toxic FN-risk & 35\% & Low toxic probability among toxic examples \\
Identity-related FN-risk & 25\% & Identity score and FN risk \\
Non-toxic FP-risk & 25\% & High toxic probability among non-toxic examples \\
High-entropy examples & 15\% & Predictive entropy \\
\bottomrule
\end{tabular}
\end{table}

The toxic false-negative-risk component targets toxic examples that the model is likely to miss. The identity-related component focuses this safety objective on identity-targeted harm when the required annotations are available. The false-positive-risk component adds non-toxic examples that the model is likely to over-flag, reducing the chance that adaptation shifts the classifier too aggressively toward toxic predictions. The entropy component adds uncertain boundary cases. If a priority group contains fewer examples than its allocated budget, the remaining slots are filled using the highest-ranked remaining examples by sample-level drift score, followed by random fallback if necessary.

\subsection{LoRA-Based Model Updating}

Model updates are performed with LoRA rather than full model fine-tuning. The base transformer is frozen except for low-rank trainable adaptation matrices and the classification head. This makes the update lightweight and suitable for repeated adaptation under deployment constraints. The same LoRA adaptation procedure is used for both hard-mix and random-balanced update baselines; the difference between the methods is the construction of the adaptation set.

\section{Experimental Setup}

\subsection{Datasets and Shift Settings}

We evaluate the framework in two settings to demonstrate its flexibility across temporal and cross-dataset moderation shifts. In the Civil Comments temporal-shift setting, the baseline model is trained on earlier Civil Comments data and evaluated on later Civil Comments data. Civil Comments includes identity-related annotations, enabling direct measurement of identity-harm drift through the \texttt{identity\_attack} signal. The reported Civil experiments use fixed-seed temporal samples with up to 5{,}000 examples per year-label group and evaluate on a 6{,}500-example 2017 target split.

In the Jigsaw-to-DynaHate cross-dataset setting, the baseline model is trained on Jigsaw toxic-comment data and evaluated on DynaHate. This setting tests a stronger shift because DynaHate contains adversarially collected hate-speech examples that differ from the source training distribution. The reported runs use 12{,}000 Jigsaw training examples, 3{,}000 Jigsaw holdout examples, 3{,}000 synthetic drift examples, 3{,}000 DynaHate adaptation-pool examples, and 3{,}000 DynaHate evaluation examples. DynaHate does not provide the same continuous identity-attack annotation as Civil Comments, so the identity-harm monitor is inactive in this setting.

\subsection{Model and Training Details}

All reported LoRA experiments use \texttt{distilroberta-base} as the base transformer classifier. LoRA is applied to the \texttt{query} and \texttt{value} modules with rank \(r=8\), \(\alpha=16\), dropout \(0.05\), and no bias adaptation. The source model is trained for 3 epochs. Drift-triggered adaptation is then performed for 1 epoch on the selected adaptation set. Training uses learning rate \(1\times 10^{-4}\), weight decay \(0.01\), warmup ratio \(0.03\), maximum sequence length 192, batch size 16, gradient accumulation 1, and FP16 precision on GPU.

\subsection{Classification Thresholds}

The toxic/non-toxic decision threshold is not fixed at 0.5 in the final reported evaluations. Instead, thresholds are calibrated to satisfy a target toxic recall of 0.85 when possible. The threshold search evaluates values from 0.05 to 0.95 in increments of 0.01. Among thresholds meeting the target recall, the selected threshold maximizes macro F1. If no threshold reaches the target recall, the selected threshold minimizes the recall shortfall and then maximizes macro F1. Baseline thresholds are calibrated on the source validation or holdout split. Adapted-model thresholds are calibrated on a held-out threshold-tuning subset of the adaptation pool. The same threshold-calibration procedure is applied to both hard-mix and random-balanced adapted models.

\subsection{Monitoring Windows and Trigger Thresholds}

For each experiment, the reference batch is the model's source-domain validation or holdout distribution, and the target batch is the shifted evaluation distribution. The monitoring windows therefore correspond to the same fixed evaluation splits used for reporting drift and model behavior. For adaptation pools, 50\% of the pool is reserved for threshold tuning and the remaining 50\% is available for sample selection.

The global Jensen-Shannon drift threshold is 0.30. The identity-attack rate threshold is 0.005 in the reported harm-aware runs, and the false-negative-risk drift threshold is 0.005. The default identity-score, mean-toxic-probability, and entropy thresholds are 0.01, 0.03, and 0.03, respectively. These thresholds are treated as operating points for the experimental protocol rather than universal constants.

\subsection{Adaptation Budget and Baselines}

Each LoRA update uses an adaptation budget of \(K=800\) examples. Under hard-mix selection, this corresponds to target allocations of 280 toxic false-negative-risk examples, 200 identity-related toxic false-negative-risk examples, 200 non-toxic false-positive-risk examples, and 120 high-entropy examples. When identity annotations are unavailable, as in DynaHate, the identity-related allocation is filled by fallback ranked examples.

We compare hard-mix adaptation with two baselines. The no-update baseline evaluates the original model on the shifted target data without adaptation. The random-balanced update baseline adapts the model using 800 randomly selected recent examples balanced across toxic and non-toxic labels. This comparison tests whether the gains come from updating on recent data generally or from risk-aware sample selection.

\subsection{Evaluation Metrics and Replication}

We evaluate model performance using four primary metrics:
\begin{itemize}
\item \textbf{Macro F1:} Measures class-balanced classification quality.
\item \textbf{Toxic recall:} Measures label-conditional recall on toxic examples.
\item \textbf{Aggregate false-negative prevalence:} Measures the proportion of all evaluation examples that are toxic but predicted as non-toxic.
\item \textbf{Accuracy:} Measures overall classification correctness.
\end{itemize}

All reported LoRA adaptation experiments are repeated over three random seeds: 13, 42, and 101. For the DynaHate robustness experiment, we additionally compute bootstrap confidence intervals by resampling pooled prediction rows with replacement and recomputing metric deltas between the no-update and adapted models.

\section{Experiments}

This section reports the experimental results for multi-monitor drift detection and selective adaptation on Civil Comments and DynaHate. Reported values are averaged across three random seeds. Adaptation methods are compared under the same adaptation budget and evaluated on the same target splits.

\subsection{Civil Comments Results}

\subsubsection{Drift Monitor Outcomes}

In the Civil Comments temporal experiment, global JS drift does not trigger under the original global threshold, but identity-harm drift does.

\begin{table}[t]
\centering
\caption{Civil Comments drift monitor outcomes.}
\label{tab:civil-drift-monitors}
\begin{tabular}{lrr}
\hline
Monitor & Value & Triggered \\
\hline
Global JS drift & 0.0775 & No \\
Identity attack rate delta & +0.0325 & Yes \\
Identity attack score delta & +0.0313 & Yes \\
Entropy delta & +0.0065 & No \\
Mean toxic probability delta & -0.0007 & No \\
Predicted toxic prevalence delta & -0.0011 & No \\
False-negative-risk delta & +0.0123 & Yes\\
\hline
\end{tabular}
\end{table}

As shown in Table~\ref{tab:civil-drift-monitors}, the Civil Comments result shows a targeted safety shift rather than broad distributional drift. Global JS drift remains low (0.0775) and does not trigger adaptation, while both identity-harm monitors trigger: identity attack rate increases by 0.0325 and mean identity attack score increases by 0.0313. In addition, false-negative-risk drift increases by 0.0123, indicating that the model is more likely to miss toxic examples under the shifted distribution. By contrast, entropy, mean toxic probability, and predicted toxic prevalence remain nearly unchanged. This supports our central claim: global drift alone is insufficient for adaptive moderation. Safety-relevant drift can emerge in harm subspaces even when the overall input distribution appears stable.

\subsubsection{Adaptation Performance}
As shown in Table~\ref{tab:civil-strategy-comparison}, the strategy comparison shows that hard-mix updating provides the strongest overall adaptation outcome. Compared with random-balanced updating, hard-mix achieves substantially higher toxic recall, increasing recall from 0.8501 to 0.8777. This indicates that hard-mix better recovers toxic-content detection, the primary safety objective in moderation. Hard-mix also achieves higher overall accuracy, improving from 0.8238 under random-balanced updating to 0.8334. Although random-balanced updating reports a lower aggregate false-negative prevalence, hard-mix provides the better tradeoff between safety recall and overall predictive performance. These results suggest that selecting high-risk and high-uncertainty examples is more effective than updating on a general balanced sample when the goal is to preserve moderation safety under drift.
\begin{table}[t]
\centering
\caption{Civil Comments update strategy comparison.}
\label{tab:civil-strategy-comparison}
\scriptsize
\setlength{\tabcolsep}{3pt}
\renewcommand{\arraystretch}{1.08}
\begin{tabular}{lcccc}
\toprule
Update method & Macro F1 & Tox. recall & FN prev. & Acc. \\
\midrule
No update & 0.8148 & 0.8453 & 0.0773 & 0.8151 \\
Random-balanced & 0.8236 & 0.8501 & 0.0499 & 0.8238 \\
Hard-mix & 0.8229 & 0.8777 & 0.0611 & 0.8334 \\
\bottomrule
\end{tabular}
\end{table}

\subsection{DynaHate Results}

\subsubsection{Drift Monitor Outcomes}

Compared with Civil Comments, DynaHate exhibits a larger cross-domain distributional shift. The global JS drift monitor increases substantially, and the model-risk monitors show substantial increases in uncertainty, predicted toxicity, and false-negative risk.

\begin{table}[t]
\centering
\caption{DynaHate drift monitor outcomes.}
\label{tab:dynahate-drift-monitors}
\begin{tabular}{lrr}
\hline
Monitor & Value & Triggered \\
\hline
Global JS drift & 0.2796 & No \\
Identity-harm drift & 0.0000 & Not available \\
Entropy delta & +0.2188 & Yes \\
Mean toxic probability delta & +0.1409 & Yes \\
Predicted toxic prevalence delta & +0.1653 & Yes \\
False-negative-risk delta & +0.1311 & Yes \\
\hline
\end{tabular}
\end{table}

As shown in Table~\ref{tab:dynahate-drift-monitors}, the DynaHate results show a substantial cross-domain shift relative to the Jigsaw training distribution. Global JS drift reaches 0.2796, below the 0.30 threshold. However, model-risk monitors increase substantially: entropy (+0.2188), mean toxic probability (+0.1409), predicted toxic prevalence (+0.1653), and false-negative-risk (+0.1311). These results indicate that the DynaHate shift affects both the input distribution and the model's safety-relevant behavior, especially its risk of missing toxic content.

\subsubsection{Adaptation Performance}

Hard-mix adaptation substantially improves toxic recall and false-negative prevalence on DynaHate.

\begin{table}[t]
\centering
\caption{DynaHate update strategy comparison.}
\label{tab:dynahate-strategy-comparison}
\scriptsize
\setlength{\tabcolsep}{3pt}
\renewcommand{\arraystretch}{1.08}
\begin{tabular}{lcccc}
\toprule
Update method & Macro F1 & Tox. recall & FN prev. & Acc. \\
\midrule
No update & 0.5328 & 0.7107 & 0.1594 & 0.5568 \\
Random-balanced & 0.5255 & 0.8171 & 0.0787 & 0.5864 \\
Hard-mix & 0.5343 & 0.8523 & 0.0813 & 0.6010 \\
\bottomrule
\end{tabular}
\end{table}

As shown in Table~\ref{tab:dynahate-strategy-comparison}, the DynaHate strategy comparison shows that both adaptation methods improve safety-relevant performance relative to the no-update baseline. Random-balanced updating increases toxic recall from 0.7107 to 0.8171 and reduces the false-negative prevalence from 0.1594 to 0.0787. Hard-mix updating produces the strongest overall result, achieving the highest macro F1 (0.5343), toxic recall (0.8523), and accuracy (0.6010). Although random-balanced updating has a slightly lower false-negative prevalence, hard-mix provides the best balance between toxic-content detection and overall predictive performance. This suggests that risk-aware sample selection is more effective than general balanced updating under the stronger DynaHate cross-domain shift.

\subsubsection{Bootstrap Confidence Intervals}

The bootstrap analysis evaluates whether the observed DynaHate improvements are stable under resampling of the evaluation predictions. We repeatedly resample the pooled predictions with replacement and recompute the metric deltas between the no-update and adapted models. As shown in Table~\ref{tab:dynahate-bootstrap-ci}, the resulting confidence intervals show that the safety gains are robust: toxic recall increases by 0.1418 with a 95\% CI of [0.1305, 0.1531], and false-negative prevalence decreases by 0.0781 with a 95\% CI of [-0.0846, -0.0717]. By contrast, macro F1 changes only slightly and its interval includes zero, suggesting that adaptation mainly improves the safety-critical ability to detect toxic content rather than broadly improving all classification metrics.

\begin{table}[t]
\centering
\caption{Bootstrap confidence intervals for DynaHate metric changes.}
\label{tab:dynahate-bootstrap-ci}
\begin{tabular}{lrr}
\hline
Metric delta & Mean & 95\% CI \\
\hline
Toxic recall & +0.1418 & [+0.1305, +0.1531] \\
False-negative prevalence & -0.0781 & [-0.0846, -0.0717] \\
Accuracy & +0.0442 & [+0.0354, +0.0531] \\
Macro F1 & +0.0017 & [-0.0081, +0.0113] \\
\hline
\end{tabular}
\end{table}

\subsection{Cross-Experiment Findings}

The updated experiments support two main findings.

\textbf{Finding 1: Multiple monitors detect safety-relevant drift that global drift alone does not capture.} The Civil Comments result shows a targeted harm shift: global JS drift remains low at 0.0775 and does not trigger, while identity attack rate drift (+0.0325), identity attack score drift (+0.0313), and false-negative-risk drift (+0.0123) all trigger. This pattern indicates that harmful behavior can change in a safety-critical subspace even when the overall text distribution appears stable. In contrast, DynaHate shows a broader cross-domain shift. Global JS drift reaches 0.2796 but remains below the operating threshold, while the model-risk monitors increase substantially in entropy (+0.2188), mean toxic probability (+0.1409), predicted toxic prevalence (+0.1653), and false-negative risk (+0.1311). Together, the two settings show why adaptive moderation should monitor both distributional drift and safety-specific model-risk signals. Civil Comments demonstrates that global drift can miss targeted harm-subspace changes, while DynaHate demonstrates that stronger cross-domain shift can affect both input distributions and model behavior.

\textbf{Finding 2: Risk-aware hard-mix updating provides a stronger adaptation tradeoff than general balanced updating.}
After drift is detected, updating the model with carefully selected samples improves safety-oriented performance. Hard-mix updating raises toxic recall to 0.8777 and accuracy to 0.8334 on Civil Comments, and raises toxic recall from 0.7107 to 0.8523 and accuracy from 0.5568 to 0.6010 on DynaHate. The DynaHate bootstrap analysis confirms that these safety gains are stable: toxic recall increases by 0.1418 with a 95\% confidence interval of [0.1305, 0.1531], and false-negative prevalence decreases by 0.0781 with a 95\% confidence interval of [-0.0846, -0.0717].

The comparison with random-balanced updating shows why sample selection matters. On Civil Comments, hard-mix achieves higher toxic recall than random-balanced updating (0.8777 vs. 0.8501) and higher accuracy (0.8334 vs. 0.8238). On DynaHate, hard-mix also outperforms random-balanced updating on toxic recall (0.8523 vs. 0.8171), macro F1 (0.5343 vs. 0.5255), and accuracy (0.6010 vs. 0.5864). Although random-balanced updating remains competitive on aggregate false-negative prevalence, hard-mix better matches our objective. It selects high-risk and high-uncertainty samples to recover safety recall while preserving overall performance.

\section{Conclusion}

This paper evaluated whether adaptive moderation can be improved by connecting drift detection to safety-aware model updating. The results show that drift in moderation is not always visible through a single global distribution metric. In the Civil Comments temporal setting, safety-relevant identity-harm signals changed even when global drift did not trigger. In the DynaHate transfer setting, global drift increased but remained below the operating threshold, while model-risk monitors increased substantially, indicating a broader and more severe safety-relevant shift. Together, these findings support the use of multiple monitors that separately track distributional change, harm-subspace change, uncertainty, toxic-risk, and false-negative risk.

The adaptation results further show that the choice of update data matters. Updating the model with hard-mix examples improved toxic recall and reduced missed toxic content, especially under the stronger DynaHate shift. Compared with random-balanced updating, hard-mix adaptation produced stronger toxic-recall and accuracy tradeoffs across the reported experiments. This suggests that adaptive moderation should not only decide when to update, but also select examples according to the failure modes that triggered the update.

This study also has limitations that suggest directions for future work. Some harm-aware monitors depend on dataset-specific annotations, such as Civil Comments identity-attack scores, and should be extended with annotation-independent harm-subspace monitors. The trigger thresholds are operational choices rather than universal constants, so future work should study calibration through validation data, risk-cost curves, or conformal monitoring. Finally, our experiments use offline benchmarks. Real-world moderation also involves delayed labels, adversarial behavior, policy changes, and human-review costs. Future work should evaluate DriftGuard in online or human-in-the-loop moderation workflows, including carefully validated model-assisted review mechanisms for prioritizing uncertain cases and generating weak safety signals\cite{yue2025real}. Future work could also study DriftGuard-style monitoring in more interactive or agentic language-system settings, where model behavior may evolve through user feedback and system actions \cite{su2026agentic}.

Overall, the study supports a safety-aware view of moderation drift. Model updates should be triggered by emerging harm and model risk, not only aggregate distribution shift. A multi-monitor trigger paired with selective hard-mix adaptation provides a practical step toward moderation systems that can respond more directly to evolving harmful behavior while limiting unnecessary or poorly targeted retraining.

\bibliographystyle{IEEEtran} 
\bibliography{main}    

\end{document}